\documentclass[10pt,twocolumn,letterpaper]{article}

\usepackage[accsupp]{axessibility}  % Improves PDF readability for those with disabilities.
\usepackage{wacv}
\usepackage{times}
\usepackage{epsfig}
\usepackage{graphicx}
\usepackage{amsmath}
\usepackage{amssymb}
\usepackage{booktabs}
\usepackage{balance}
\usepackage{algorithm}
\usepackage[noend]{algpseudocode}
\usepackage{multirow}
\usepackage{booktabs}
\usepackage[pagebackref=true,breaklinks=true,colorlinks,bookmarks=false]{hyperref}
\usepackage{subcaption}
\usepackage{tablefootnote}
\usepackage[dvipsnames]{xcolor}

\def\wacvPaperID{166} % Enter the WACV Paper ID here

\wacvfinalcopy % *** Uncomment this line for the final submission

\ifwacvfinal
\usepackage[breaklinks=true,bookmarks=false]{hyperref}
\else
\usepackage[pagebackref=true,breaklinks=true,colorlinks,bookmarks=false]{hyperref}
\fi

\usepackage{caption}
\usepackage{subcaption}

\begin{document}

%%%%%%%%% TITLE
\title{Federated Domain Generalization for Image Recognition via \\Cross-Client Style Transfer}

\author{
% Junming Chen$^1$\thanks{Joint first authors. Junming proposed the idea, wrote the main paper, and implemented the proposed method and privacy experiment. Meirui implemented other methods for comparison and wrote part of the paper.} \quad Meirui Jiang$^2$\footnotemark[1]  \quad Qi Dou$^2$ \quad Qifeng Chen$^1$\\
Junming Chen$^1$\thanks{Joint first authors.} \quad Meirui Jiang$^2$\footnotemark[1]  \quad Qi Dou$^2$ \quad Qifeng Chen$^1$\\
$^1$HKUST \quad $^2$CUHK\\
{\tt \small \{jchenfo, cqf\}@ust.hk \quad \{mrjiang, qdou\}@cse.cuhk.edu.hk}
}

\maketitle
\thispagestyle{empty}

%%%%%%%%% ABSTRACT
\begin{abstract}
   Domain generalization (DG) has been a hot topic in image recognition, with a goal to train a general model that can perform well on unseen domains. 
   Recently, federated learning (FL), an emerging machine learning paradigm to train a global model from multiple decentralized clients without compromising data privacy, has brought new challenges and possibilities to DG.
   In the FL scenario, many existing state-of-the-art (SOTA) DG methods become ineffective because they require the centralization of data from different domains during training. 
   In this paper, we propose a novel domain generalization method for image recognition under federated learning through cross-client style transfer (CCST) without exchanging data samples. 
   Our CCST method can lead to more uniform distributions of source clients, and make each local model learn to fit the image styles of all the clients to avoid the different model biases. 
   Two types of style (single image style and overall domain style) with corresponding mechanisms are proposed to be chosen according to different scenarios. Our style representation is exceptionally lightweight and can hardly be used to reconstruct the dataset. The level of diversity is also flexible to be controlled with a hyper-parameter.
   Our method outperforms recent SOTA DG methods on two DG benchmarks (PACS, OfficeHome) and a large-scale medical image dataset (Camelyon17) in the FL setting. Last but not least, our method is orthogonal to many classic DG methods, achieving additive performance by combined utilization. Our code is available at: \href{https://chenjunming.ml/proj/CCST}{https://chenjunming.ml/proj/CCST}. 
\end{abstract}

%%%%%%%%% BODY TEXT

\section{Introduction}
\label{sec:intro}
\begin{figure*}[t]
\centering
\includegraphics[width=0.8\textwidth]{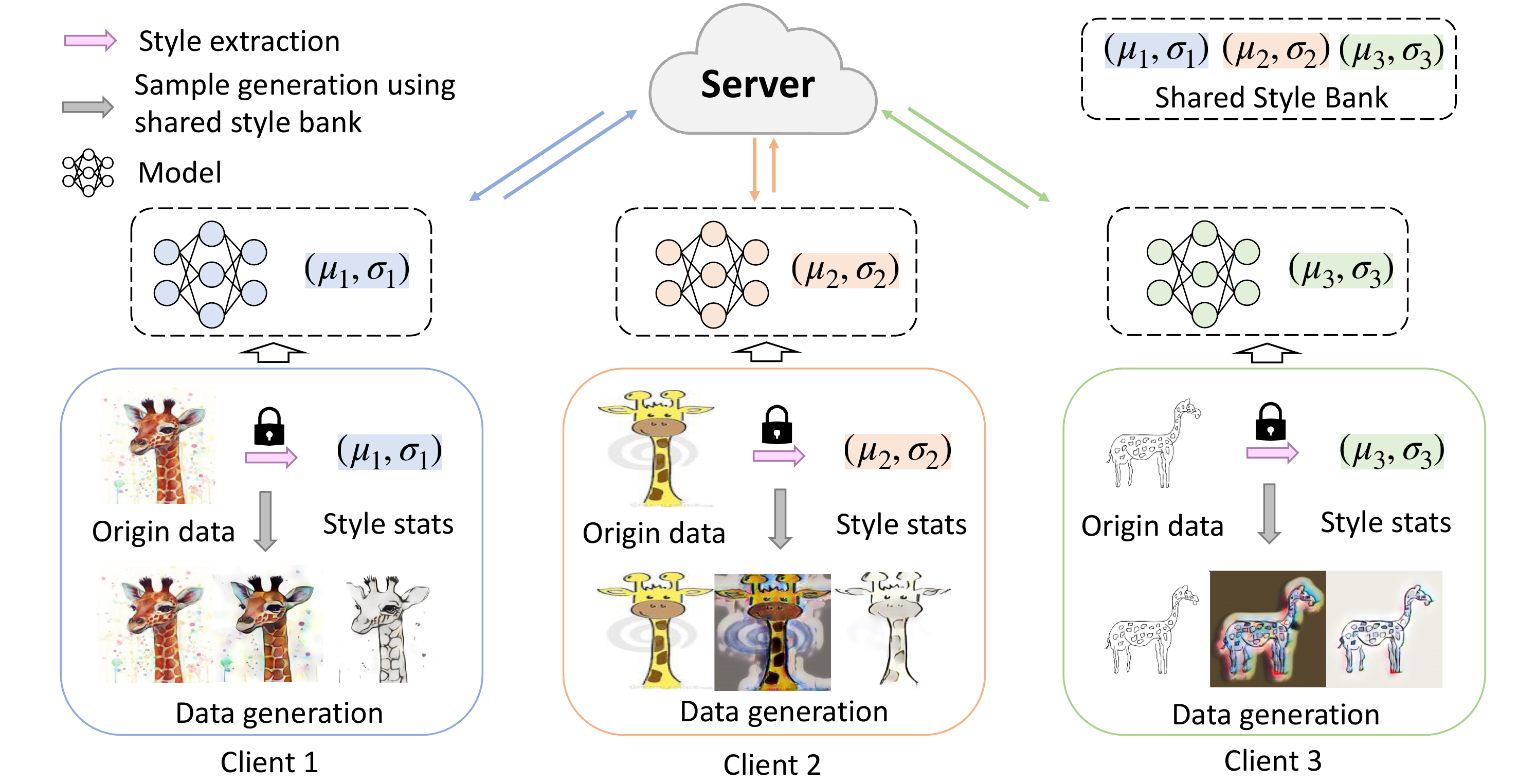}
\vspace{-2mm}
\caption{Overview of our framework with style transfer across source clients using three different source styles on the PACS dataset. We augment each source client data with styles of other two source clients.}
\label{fig:fl_framework}
\end{figure*}
Federated learning (FL) aims to train a machine learning model on multiple decentralized local clients without explicitly exchanging data samples. This emerging technique has triggered increasing research interest in recent years, owing to its significant applications in many real-world scenarios such as finance, healthcare, and edge computing \cite{kairouz2019advances}. The paradigm works in a way that each local client (e.g., hospital) learns from their local data and only aggregates the model parameters at a specific frequency on the central server to yield a global model.
%All the data samples are kept within each local client in FL to ensure privacy.

One of the biggest challenges in FL is tackling the non-identically and independently distributed (non-IID) data across different clients. Although much progress has been made on addressing non-IID issues in FL~\cite{li2021fedbn,li2018federated}, most of them only focus on improving the performance of internal clients. Few papers focus on domain generalization in FL, which is a crucial scenario considering the model generalization ability on a new client with unseen data distribution. For example, it is important that a federated trained disease diagnose model by multiple hospitals can be directly utilized by other new hospitals with a high accuracy, especially when they have few annotated data to train a good model. DG aims to improve the test performance on unseen target domains with the model trained on multi-source data. A prior work FedDG \cite{liu2021feddg} proposes to exchange the amplitude information in frequency domain cross clients and utilize episodic learning to improve the performance further. However, they are specific to medical image segmentation tasks and consider the distribution shift across medical imaging protocols, which remains unexplored for larger domain gaps in the wild. In contrast, we aim to improve the model generalization ability for image recognition tasks, 
and our method is able to handle domain shifts from small (cross-site medical images) to more significant ones like photos and sketches in the PACS dataset.
% and our method not only performs better than FedDG on medical images, but also is able to handle more significant domain shifts like photos and sketches in the PACS dataset.

The FL scenario poses particular and new challenges to DG: regarding each client as a domain with a specific style, the data from each domain cannot be put together during training, which violates the implicit requirement of many DG methods. For example, meta-learning \cite{li2019episodic} and adversarial domain invariant feature learning \cite{Li2018} both require access to all the source domains at the same time, which is not directly applicable in federated learning. In addition,  straightforward aggregating the parameters of local models may lead to a sub-optimal global model because the local models are biased to different client styles. To solve those problems, we propose a data-level cross-domain style transfer (CCST) method that augments the data by using other source domain styles with the style transfer technique. In this way, each client will have styles of all the other source clients, and thus all the local models will have the same goal to fit images with all the source styles, which avoids the different local model biases that may compromise the global model performance. Moreover, CCST is orthogonal to other DG methods, and thus existing methods on centralized DG can also benefit from a further accuracy boost.

Our CCST method for federated domain generalization is general and compatible with any style transfer method that satisfies two requirements:
First, the style information in the style transfer algorithm cannot be utilized to reconstruct the dataset; 
Second, this style transfer method should be an arbitrary style transfer per model method, which means the style transfer model should be ready to transfer a content image to arbitrary styles. Since there can be many clients in federated learning, the style transfer model should better have the ability to transfer all those styles without retraining. Otherwise, the deployment cost will significantly increase since each client has to store various models locally for different styles and even require retraining for unseen styles.
% if each style need to have a style transfer model, then each client has to have all of those models in their locally, which increase the deployment cost. 
In our paper, we choose AdaIN \cite{Huang_2017_ICCV}, an effective real-time arbitrary style transfer model to demonstrate the effectiveness of our CCST framework. The style information used in AdaIN is the moments (i.e., mean and variance) of each pixel-level feature channel at a specific VGG layer, which are extremely lightweight (two 512-dimensional vectors) and do not contain spatial structural information about the image content. Therefore, such style information is efficient to be shared across clients and can hardly lead to the reconstruction of the dataset. Further analysis could be found in Section~\ref{sec:discussion}.  

The overall framework of our method is shown in Figure \ref{fig:fl_framework}. Each client is regarded as a domain with a domain-specific style. Before training the image recognition model, we first compute the style information of images in each source client. We design two types of styles that can be shared: single image style and overall domain style, which will be illustrated in detail in Section \ref{sec:method}. Then the source clients will upload their style information to the global server and share them with all the source clients, which we call style bank. Each source client utilizes the shared style bank to perform style transfer on their local data, during which a hyperparameter K is introduced to control the diversity level of our CCST process. Federated training will begin after each source client finishes data augmentation, and then the trained model will be directly tested on the unseen target client.
Our contributions are summarized below:

% \begin{itemize}
    % \renewcommand{\labelitemi}{\textbullet}
    % \item 
    \textbf{(a)} We propose a simple yet effective framework named cross-client style transfer (CCST).
    % , which is the first method of domain generalization for image recognition under the horizontal federated learning setting to the best of our knowledge. 
    Our approach achieves new state-of-the-art generalization performance in FL setting on two standard DG benchmarks (PACS~\cite{PACS}, OfficeHome~\cite{officehome}) and a large-scale medical image dataset (Camelyon17~\cite{camelyon17}).
    % \item Our method can handle larger domain gap compared with other DG methods for FL, because the AdaIn in our framework can perform style transfer with large domain shift. 
    % \item 
    \textbf{(b)}  Two types of styles with corresponding sharing mechanisms are proposed, named \textit{overall domain style} and \textit{single image style}, which can be chosen according to different circumstances. The diversity level of our method is also flexible to be adjusted.
    % \item 
    \textbf{(c)} The proposed method is orthogonal to many other SOTA DG methods. Therefore, our method can be readily applied to those DG methods to have a further performance boost. We also study the effectiveness of several SOTA DG methods when they are applied in the FL setting for image recognition.
    \textbf{(d)} We give an intuitive (Section \ref{sec:discussion}) and experimental analysis (Section~\ref{sec:privacy}) on the privacy-preserving performance of our style vectors to demonstrate that one can hardly reconstruct the original images merely from the style vectors using the generator from a SOTA GAN \cite{liu2021towards} in FL setting.
% \end{itemize}

\section{Related Work}
% \subsection{Federated Learning with Non-IID Data}
% \textbf{Federated learning with non-IID data.}
% Non-IID data distribution across clients can arise in various ways in FL, including label distribution skew, feature distribution skew, concept drift, concept shift, and quantity skew \cite{kairouz2019advances}. 
% Among them, concept drift and feature distribution skew are less explored in the literature. In this case, different clients have the same label space but different feature distribution, e.g., real dogs in photos and those drew by hand have very different feature distribution. To solve this problem, Reisizadeh et al. \cite{reisizadeh2020robust} assume data follows an affine distribution shift and introduces a fast and efficient optimization method to tackle the problem. Recently, FedBN \cite{li2021fedbn} is proposed for the setting where data of each client have the same labels but different features. FedBN excludes the Batch Normalization (BN) layer from aggregating step and has a better performance than FedProx and FedAvg. However, those methods only focus on improving the performance of internal clients without considering the performance of the global model on newly joined unseen clients. In our scenario, all the clients also have different feature distributions including the unseen client, but we mainly focus on the model generalization performance on the unseen client.

% \subsection{Domain Generalization}
\textbf{Domain generalization.}
 Domain generalization is a popular research field that aims to learn a model from multiple source domains such that the model can generalize on the unseen target domain. Many works are proposed towards solving the domain shifts from various directions under the centralized data setting. Those methods can be divided into three categories~\cite{wang2021generalizing}, including manipulating data to enrich data diversity~\cite{jackson2019style,xu2021fourier,shankar2018generalizing,zhou2021domain}, learning domain-invariant representations or disentangling domain-shared and specific features to enhance the generalization ability of model~\cite{arjovsky2019invariant,piratla2020efficient,carlucci2019domain,zhao2020domain} and exploiting general learning strategies to promote generalizing capability~\cite{li2019episodic,huang2020self,dou2019domain,du2020learning}.

However, many of these methods require centralized data of different domains, violating the local data preservation in federated learning. Specifically, access for more than one domain is needed to augment data or generate new data in~\cite{shankar2018generalizing,jackson2019style}, domain invariant representation learning or decomposing features is performed under the comparison across domains~\cite{arjovsky2019invariant,piratla2020efficient,zhao2020domain} and some learning strategy based methods utilize extra one domain for meta-update~\cite{li2019episodic,dou2019domain,du2020learning}. Nevertheless, some methods do not explicitly require centralized domains or can be adapted into federated learning with minor changes. For example, MixStyle~\cite{zhou2021domain} can optionally conduct the style randomization in a single domain to augment data; \cite{xu2021fourier} uses Fourier transformation to augmentation that is free of sharing data; JiGen~\cite{carlucci2019domain} proposes a self-supervised task to enhance representation capability; RSC~\cite{huang2020self} designs a learning strategy based on gradient operations without explicit multi-domain requirements.

\textbf{Federated / decentralized domain generalization.} Despite many works on centralized domain generalization and tackling non-IID issues in FL, there are few works addressing the DG problem in FL. 
FedDG~\cite{liu2021feddg} exchanges the amplitude information across images from different clients and utilizes episodic learning to improve performance further. However, it only focuses on the segmentation task with superficial domain shift in data, and its performance on image recognition with larger domain shift remains unexplored.  
COPA~\cite{Wu_2021_ICCV} propose only aggregating the weights for domain-invariant feature extractor and maintaining an assemble of domain-specific classifier heads to tackle the decentralized DG. However, since COPA has to share classifier heads of all the clients locally and globally, it may lead to privacy issues, heavier communication, and higher test-time inference cost. 

% On the other hand, this method is in aggregation-level, which is compatible with our data-level CCST method. 

\textbf{Neural style transfer.}
\label{style_transfer}
Neural style transfer (NST) aims to transfer the style of an image to another content image with its semantic structure reserved. The development of NST has roughly gone through three stages: per-style-per-model (PSPM), multiple-style-per-model (MSPM) and arbitrary-style-per-model (ASPM) methods \cite{StyleTransferReview}. PSPM methods \cite{Gatys_2016_CVPR,Johnson2016PerceptualLF,Ulyanov2016TextureNF} can only transfer a single style for each trained model. MSPM methods \cite{dumoulin2016learned,stylebank,Zhang2018MultistyleGN,Li2017DiversIfiedTS}  are able to transfer multiple styles with a single trained model. However, PSPM and MSPM are expensive to deploy when too many styles are required to be transferred in our setting. ASPM \cite{chen2016fast,huang2017adain,Ghiasi2017,Li2017} can transfer arbitrary styles to any content images and is often faster than PSPM and MSPM, which is more suitable for our scenario. 
%Among those methods, AdaIN \cite{huang2017adain} is the first real-time arbitrary style transfer method, and is also the neatest and fastest real-time ASPM method. It performs de-stylization by normalizing the VGG feature with its own moments and then stylizes itself by affine transformation with the moments of the style image feature.

The first ASPM method is proposed by Chen and Schmidt \cite{chen2016fast}, but it cannot achieve real-time. AdaIN \cite{huang2017adain} is the first real-time arbitrary style transfer method, which utilizes the channel-wise mean and variance as style information. It performs de-stylization by normalizing the VGG feature with its own style and then stylizes itself by affine transformation with the mean and variance of the style image feature. Another real-time ASPM method \cite{Ghiasi2017} is a follow-up work of CIN \cite{Dumoulin2017}. They change the MSPM method CIN into an ASPM method by predicting the affine transformation parameters for each style image through another style prediction network. However, the level of style-content disentanglement of the predicted style vector remains unknown, which may have privacy issue in FL setting. Later, Li et al. \cite{Li2017} propose a universal style-learning free ASPM method, which utilizes ZCA whitening transform for de-stylization and coloring transform for style transfer. However, this method is much slower than previous methods in practice. Therefore, we choose the neatest and efficient real-time ASPM method AdaIN as our style transfer model in our framework.

\section{Method}
\label{sec:method}
% Introduce why we use this method for DG, the high-level intuition. high-level explanation why this works.
The core idea of our method is to let the distributed clients have as similar data distribution as possible by introducing styles of other clients into each of them via cross-client style transfer without dataset leakage. Figure~\ref{fig:hist} shows the data distribution before and after our CCST method. In this way, we can make the trained local models learn to fit all the source client styles and avoid aggregating the local models biased to different styles. As a result, each client can be regarded as a deep-all \cite{carlucci2019domain} setting, and the local models will have the same goal to fit styles from all the source clients. We propose two types of styles that can be chosen to transfer: one is overall domain style, the other is single image style.  In the following sections, we will introduce our reorganized style transfer framework and the process of cross-client style transfer.

%\subsubsection{Preliminaries}
Our CCST method for federated domain generalization is general and compatible with any style transfer method that satisfies two requirements:
For a style transfer model to be utilized in our general CCST framework, at least two requirements should be satisfied: 1) The style information shared among clients cannot be utilized to reconstruct the dataset; 2) The style transfer method should be a real-time arbitrary style transfer model to allow efficient and straightforward style transfer. AdaIN~\cite{huang2017adain} is the first real-time arbitrary style transfer model, and it perfectly satisfies both requirements. Moreover, it is extremely difficult to recover the dataset only from the style information utilized in AdaIN. We give an intuitive analysis in Section~\ref{sec:discussion} about the privacy issue. Therefore, we choose AdaIN as our style transfer model to demonstrate the effectiveness of our CCST framework.
Formally, for content image $I_c$ and style image $I_s$, the corresponding VGG features $F_c$ and $F_s$ are:
\begin{equation}
 F_c = \Phi(I_c), \ \ F_s = \Phi(I_s),
\end{equation}
where $\Phi$ is the VGG encoder. 

The AdaIN module takes the VGG features $F_c$ and $F_s$ of content and style images as input, which first normalize (de-stylization) with moments of $F_c$ and then perform an affine transformation (stylize) with the moments of $F_s$. Formally, 
\begin{equation}
{AdaIN}(F_{c}, F_{s})=\sigma(F_{s})\left(\frac{F_{c}-\mu(F_{c})}{\sigma(F_{c})}\right)+\mu(F_{s}),
\end{equation}
where $\mu(\cdot)$ and $\sigma(\cdot)$ compute channel-wise mean and standard variance of image features. Assume $F_c$ have the shape of $x \times h \times w$ given channel dimension $x$ and feature map resolution $h \times w$, then $\mu(F_c)$ and $\sigma(F_c)$ have the same shape of $x \times 1$.  Normally, $x=512$ in our work.

Then, the transferred feature $F_{c \leftarrow s}$ derived from ${AdaIN}(F_{c}, F_{s})$, is passed to a decoder $\Psi$ to generate the final stylized image $I_{c \leftarrow s}$:
\begin{equation}
I_{c \leftarrow s} = \Psi({AdaIN}(F_{c}, F_{s})).
\end{equation}

\begin{figure}[t]
\centering
\includegraphics[width=0.98\columnwidth]{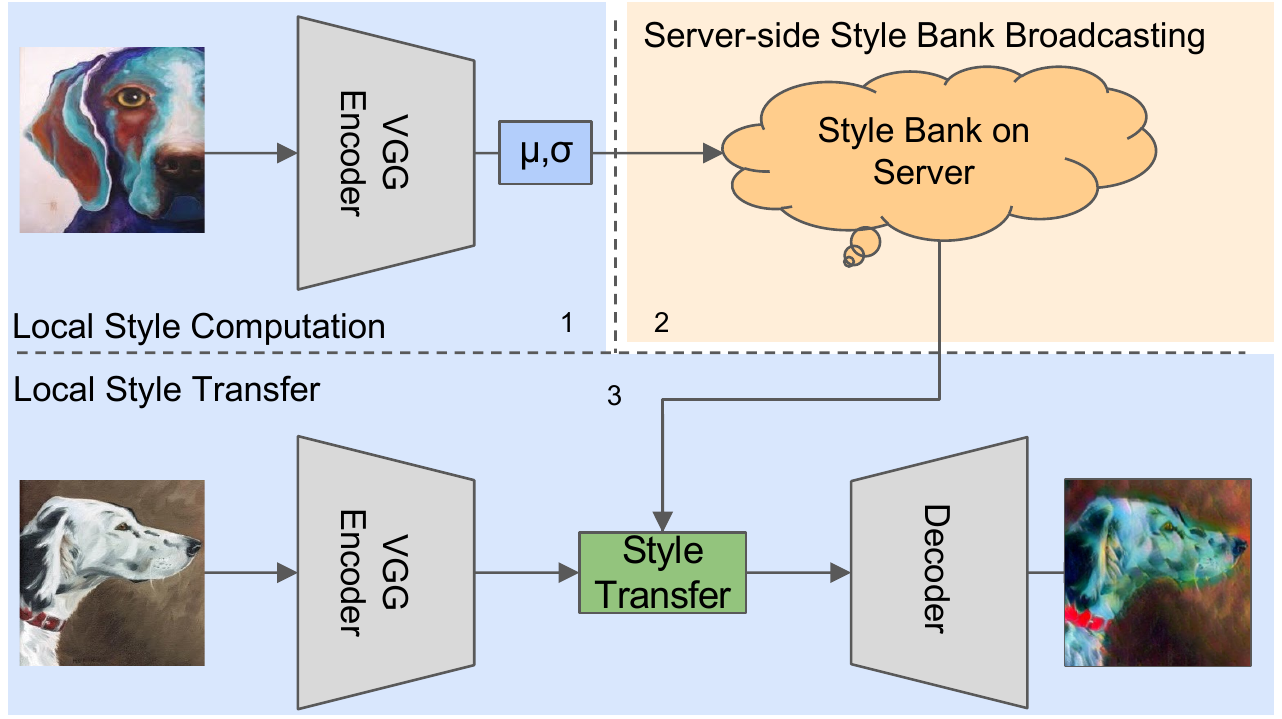} 
\vspace{-2mm}
\caption{The re-organized AdaIN \cite{huang2017adain} framework utilized for cross-client style transfer in federated learning. The VGG encoder is shared between the style extraction and image generation stage. Dash lines separate the three stages of our method: 1. Local style computation; 2. Server-side style bank broadcasting; 3. Local style transfer.}
\label{fig:adain}
\end{figure}

\subsection{Preliminaries}
As shown in Figure \ref{fig:adain}, we divide the workflow of AdaIN into two parts.
% \begin{enumerate}
    % \item 
    The first part is the style extractor $SE$ with style image as input. Denote the style as $S$, then:
    \begin{equation}
     S = SE(I_s) = (S_{\mu}, S_{\sigma})= (\mu(\Phi(I_s)), \sigma(\Phi(I_s)) ).
    \label{eq:SE}
    \end{equation}
    
    % \item 
    The second part is an image generator $G$ with content image and style vector as input:
    \begin{equation}
     I_{c \leftarrow s}  = G(I_c, S) = \Psi( S_{\sigma} \left(\frac{\Phi(I_c)-\mu(\Phi(I_c))}{\sigma(\Phi(I_c))}\right)+S_{\mu}) .
    \label{eq:G}
    \end{equation}
% \end{enumerate}
% The style extractor and image generator share the same encoder $\Phi$.

\subsection{Cross-Client Style Transfer}

We regard each client in federated learning as a domain. The workflow of the data augmentation process is shown in Algorithm \ref{data_aug}. Assume there are $N$ clients $\{C_1, C_2, ..., C_N\}$ and there is a central server. Note that we will introduce two types of styles and corresponding mechanisms in parallel in the following illustration.
As shown in Figure \ref{fig:adain}, our method has three stages:
\subsubsection{Local style computation and sharing.} 
In the beginning, each client needs to compute their style and upload them to the global server. Two types of styles can be chosen to share across clients:

% \begin{itemize}
    % \renewcommand{\labelitemi}{\textbullet}
    % \item 
    \textbf{Single image style.} Image style is calculated as the pixel-level channel-wise mean and standard variance of the VGG feature of image. Formally, for a randomly chosen image with index $i$ and VGG feature $F_i ^ {C_n}$ at client $C_n$, the single image style $S _ {single(i)} ^ {C_n}$ would be:
    \begin{equation}
     S _ {single(i)} ^ {C_n} = (\mu(F_i ^ {C_n}), \sigma(F_i ^ {C_n})).
    \end{equation}
    
    If single image styles are utilized for style transfer, multiple styles of different images should be uploaded to the server for this client to avoid single image bias and increase diversity, which forms a local image style bank $S ^ {C_n} _ {bank}$. Formally, denote randomly selected $J$ image styles to be uploaded by $C_n$ as
    \begin{equation}
     S ^ {C_n} _ {bank} = \{S _ {single(i_1)} ^ {C_n}, \hdots, S _ {single(i_J)} ^ {C_n}  \},
    \end{equation}
    where $\{i_1,\hdots,i_J\}$ are randomly sampled image indices from client $C_n$. Sharing single image styles consumes relatively low computation but can lead to high communication costs for uploading multiple styles.
    
    % \item 
    \textbf{Overall domain style.} Domain style is the domain-level channel wise mean and standard variance, which considers all the images (pixels) in a client. Formally, assume client $C_n$ has $M$ training images with corresponding VGG features $\{F_1 ^ {C_n}, F_2 ^ {C_n}, ..., F_M ^ {C_n}\}$, the overall style  $S _ {overall} ^ {C_n}$ of this client is:
    \begin{equation}
    \begin{aligned}
     &S _ {overall} ^ {C_n} = (\mu(F_{all} ^ {C_n}), \sigma(F_{all} ^ {C_n})), \\
    % \end{equation}
    % \begin{equation}
     &F_{all} ^ {C_n} = Stack(F_1 ^ {C_n}, F_2 ^ {C_n}, ..., F_M ^ {C_n}).
     \end{aligned}
    \end{equation}
    The computation cost of the overall domain style is relatively high compared to just computing several single image styles. However, since each domain only has one domain style $S _ {overall} ^ {C_n}$, choosing to upload the overall domain style to the server is more communication efficient. 
    
    Besides, the overall domain style can represent the domain style more robustly, while single image styles give more diversity and randomness to the style bank.
% \end{itemize}

\subsubsection{Server-side style bank broadcasting.}
When the server receives all the styles of each client, it will concatenate all the styles as a style bank $\mathbf{B}$ and broadcast it back to all the clients. In the two different style sharing modes, the style bank will also be different:
\begin{itemize}
\renewcommand{\labelitemi}{\textbullet}
    \item Style bank $\mathbf{B} _ {single}$ for single image styles: 
    % from each client:
    \begin{equation}
     \mathbf{B} _ {single} = \{ S ^ {C_n} _ {bank} | n = 1,2,...N \}.
    \end{equation}
    \item Style bank $\mathbf{B} _ {overall}$ for overall domain style:
    % from each client:
    \begin{equation}
     \mathbf{B} _ {overall} = \{ S _ {overall} ^ {C_n} | n = 1,2,...,N \}.
    \end{equation}
\end{itemize}

Similarly, here $\mathbf{B} _ {single}$ cost more memory than $\mathbf{B} _ {overall}$. Therefore, the latter one is more communication-friendly.

\begin{algorithm}[t]
\caption{Local Cross-Client Style Transfer at Client $C_n$}
\label{alg:algorithm}
\textbf{Input}: Training image set $\mathbf{I} ^ {C_n}$, global style bank $\mathbf{B}$. \\
\textbf{Parameter}: Augmentation level $K$, style type $T$. \\
\textbf{Output}: Augmented dataset $\mathbf{D}  ^ {C_n}$
\begin{algorithmic}[1] %[1] enables line numbers
    \State $\mathbf{D}  ^ {C_n} = [\,]$ \Comment{Augmented dataset}
    \For{$i = 1, 2, ..., m$} \Comment{$m=size(\mathbf{I})$}
        \State $S = random.choice(\mathbf{B}, K)$ 
        \For{$S ^ {C_n}$ in $S$}
            \If{$C_n$ is current client}
                \State $\mathbf{D}  ^ {C_n}.append(I_i)$
            \ElsIf{$T$ is single mode}
                %\State $S ^ {C_n} _ {single} = random.choice(S ^ {C_n}, 1)$
                %\State $I_{i,n} ' = $ \Comment{Eq. \ref{eq:G}}
                \State $\mathbf{D}  ^ {C_n}.append(G(I_i, random.choice(S ^ {C_n}, 1)))$\Comment{Eq. \ref{eq:G}}
            \ElsIf{$T$ is overall mode}
                \State $\mathbf{D}  ^ {C_n}.append(G(I_i, S ^ {C_n}))$
            \EndIf
        \EndFor
    \EndFor
    \State \textbf{return} $\mathbf{D}  ^ {C_n}$
\end{algorithmic}
\label{data_aug}
\end{algorithm}

\subsubsection{Local style transfer.} 
When client $C_n$ receives the style bank $\mathbf{B}$, the local data can be augmented by transferring styles in $\mathbf{B}$ to existing images, which introduces styles of other domains into this client.
A hyperparameter $K \in \{1,2,...,N\}$ called \textit{augmentation level}, is set to choose $K$ styles from style bank $\mathbf{B}$ for the augmentation of each image, indicating the diversity of final augmented data set. Suppose the size of the original dataset is $d$, then after cross-client style transfer, the size of the augmented data set will become $d \times K$. 

The workflow for local style transfer with two choices of styles is illustrated in Algorithm \ref{data_aug}. First, for each image $I$ in client $C_n$, $K$ random domains are selected, and each selected domain should have one style vector $S$ to be input into image generator $G$. If transfer the overall domain styles, simply choose the corresponding style $S_{overall}$ from $\mathbf{B} _ {overall}$; otherwise if transfer single images styles, one style $S_{single}$ will be randomly chosen from $S ^ {C_n} _ {bank}$ as the style of the selected domain in $\mathbf{B} _ {single}$. In both style modes, if the domain itself is chosen, this image will be directly put into the augmented data set.

% \subsection{Test-time adaptation based on cross client style transfer}

\begin{table*}[t]
\centering
\caption{Accuracy comparison of image recognition on the PACS and Office-Home dataset, each single letter column represents an unseen target client. Our CCST with the overall domain style (K=3) outperforms other methods. We use FedAvg as our base FL framework. Jigen, RSC, and Mixstyle are applied within each client. The backbone networks utilized in PACS and Office-Home are ImageNet-pretrained ResNet50 and ResNet18 respectively. \textcolor{Gray}{(\dag: Since COPA did not release their code, we copy the results from their paper here. But it cannot directly compare with our results due to the setting difference.)}}
\vspace{-2mm}
\resizebox{\textwidth}{!}{
\begin{tabular}{c@{\hspace{3mm}}ccccc|ccccc|c}
\toprule
\multirow{4}{*}{Method} & \multicolumn{5}{c}{PACS}                                & \multicolumn{5}{c}{Office-Home}                         & \multirow{4}{*}{Avg.} \\ \cmidrule(lr){2-11} 
% & \multicolumn{5}{c}{ResNet-50 (pre-trained on ImageNet)} & \multicolumn{5}{c}{ResNet-18 (pre-trained on ImageNet)} &        \\\cmidrule(lr){2-11}
& \multicolumn{1}{c}{P}         & A         & C         & S         & \multicolumn{1}{c}{Avg.}     & A         & C         & P        & R        & \multicolumn{1}{c}{Avg.}       &         \\ \midrule
FedAvg (AISTATS'17)~\cite{mcmahan2017communication}& 95.51& 82.23& 78.20& 73.56& 82.37 
& 60.08 & 45.59 & 69.48& 72.82& 61.99&72.18\\
Jigen (CVPR'19)~\cite{carlucci2019domain}& 95.99& 84.72& 77.09& 72.16  & {82.49} 
&  60.29 & 46.16 & 69.26 & 72.59& 62.07&72.28\\
RSC (ECCV'20)~\cite{huang2020self}& 95.21& 83.15& 78.24& 74.62&{82.81}
& 58.23 & 46.05  & 70.27 & \textbf{73.39}&61.99&72.40\\
MixStyle (ICLR'21)~\cite{zhou2021domain}& 95.93& 85.99& \textbf{80.03}& 75.46 & {84.35}
& 58.44 & \textbf{50.29} & {70.61} & 70.64 & 62.49&73.42\\
FedDG (CVPR'21)~\cite{liu2021feddg}& 96.23& 83.94& 79.27& 73.30& {83.19} 
 & \textbf{60.70} & 45.82 & {71.51} & {73.05}& 62.77&72.98\\
 \textcolor{Gray}{COPA-Res18\dag\ (ICCV'21)~\cite{Wu_2021_ICCV}}& \textcolor{Gray}{94.60} & \textcolor{Gray}{83.30}& \textcolor{Gray}{79.80}& \textcolor{Gray}{82.50}& \textcolor{Gray}{85.10} & \textcolor{Gray}{59.40}& \textcolor{Gray}{55.10}& \textcolor{Gray}{74.80}& \textcolor{Gray}{75.00}& \textcolor{Gray}{66.10} & \textcolor{Gray}{75.60}\\
CCST (Overall, K=3) & \textbf{96.65}& \textbf{88.33}& 78.20 & \textbf{82.90} & \textbf{86.52}
 & {59.05} & {50.06} & \textbf{72.97}& {71.67}& \textbf{63.56} & \textbf{75.04} \\ \bottomrule
\end{tabular}
}
\label{table:all_results}
\end{table*}

\section{Experiments}
\subsection{Datasets}
We evaluate our method on two standard domain generalization datasets (PACS~\cite{li2017deeper}, Office-Home~\cite{venkateswara2017deep}) that consist of various image styles as domains and a real-world medical image dataset (Camelyon17~\cite{camelyon17}). Specifically, PACS is a 7-class image recognition benchmark including 9,991 images with four different image style domains, including photo, art, cartoon, and sketch. Office-Home is another image recognition dataset that includes 15,588 images of 65 classes from four different domains (art, clipart, product, and real-world). Camelyon17 is a public tumor classification dataset, which has histology images from 5 hospitals. 
% Notably, compared with PACS, the domain shift of Office-Home is smaller. In the Office-Home dataset, the product images are collected from vendor websites mainly with white background, while the real-world represents object images collected with a regular camera, thereby these two domains share less discrepancy in image styles. 

\subsection{Experimental Settings}
\textbf{Experiment setup.} We take each domain as a single client and conduct the leave-one-domain-out experiments on PACS and Office-Home datasets. Specifically, we select one client as the target test domain and train our model on the other clients. For the medical dataset, following the setting of source/target domains in literature~\cite{camelyon17,koh2021wilds}, we apply the leave-one-domain-out setting to hospital 4 and hospital 5. For the PACS dataset, we follow the JiGen~\cite{carlucci2019domain} to split 90\% data of each client as the training set and 10\% of that as the validation set for source clients, while for unseen target clients, the entire data is used for testing. For OfficeHome and Camelyon17, which have more data samples, the ratio between train and validation set is 4:1 for each source client, and 20\% data is utilized as the test set on the unseen target client. We compare our method with \textbf{FedDG}~\cite{liu2021feddg}, which aims to solve DG problems in federated learning for medical image segmentation. We also test the performance of three centralized DG methods under FL setting (FedAvg), including \textbf{JiGen}~\cite{carlucci2019domain}, \textbf{RSC}~\cite{huang2020self} and \textbf{MixStyle}~\cite{zhou2021domain}. For \textbf{COPA}, due to its re-designed layers of ResNet18 and unknown train-validate-test split, we copy the results for reference only. We regard every single client as a centralized dataset and apply these methods locally in FL. We report the test accuracy on each unseen client by choosing the best validation model.
% and also report the average results to make an overall comparison.
\\\textbf{Implementation details.} We utilize the pre-trained AdaIN~\cite{huang2017adain} to perform style transfer. Following \cite{huang2020self}, we choose ResNet~\cite{he2016deep} pre-trained on ImageNet as our backbone for PACS and Office-Home datasets. For the Camelyon17 dataset, we follow \cite{harmofl} to use the DenseNet121~\cite{huang2017densely}. We use FedAvg~\cite{mcmahan2017communication} as our FL framework and train the model using SGD optimizer with $1e^{-3}$ learning rate for 500 communication rounds with one local update epoch on the PACS and Office-Home dataset. For Camelyon17, we train 100 communication rounds considering its large data amount.
The JiGen and RSC can be directly integrated into the FedAvg without further modifications. We adapt the MixStyle into an intra-client version that shuffles styles inside each batch of data to fit the federated setting. 
All hyper-parameters of compared methods are chosen based on corresponding papers. We follow the standard training procedure of FedAvg~\cite{mcmahan2017communication} for federated training.
The value of M in  is $\lceil dataset\_size / 32\rceil$ in our experiment. 
The framework is implemented with PyTorch and is trained on a single NVIDIA RTX 2080 Ti GPU.

\begin{figure*}[t]
     \centering
     \begin{subfigure}[b]{0.49\textwidth}
         \centering
         \includegraphics[width=0.9\textwidth]{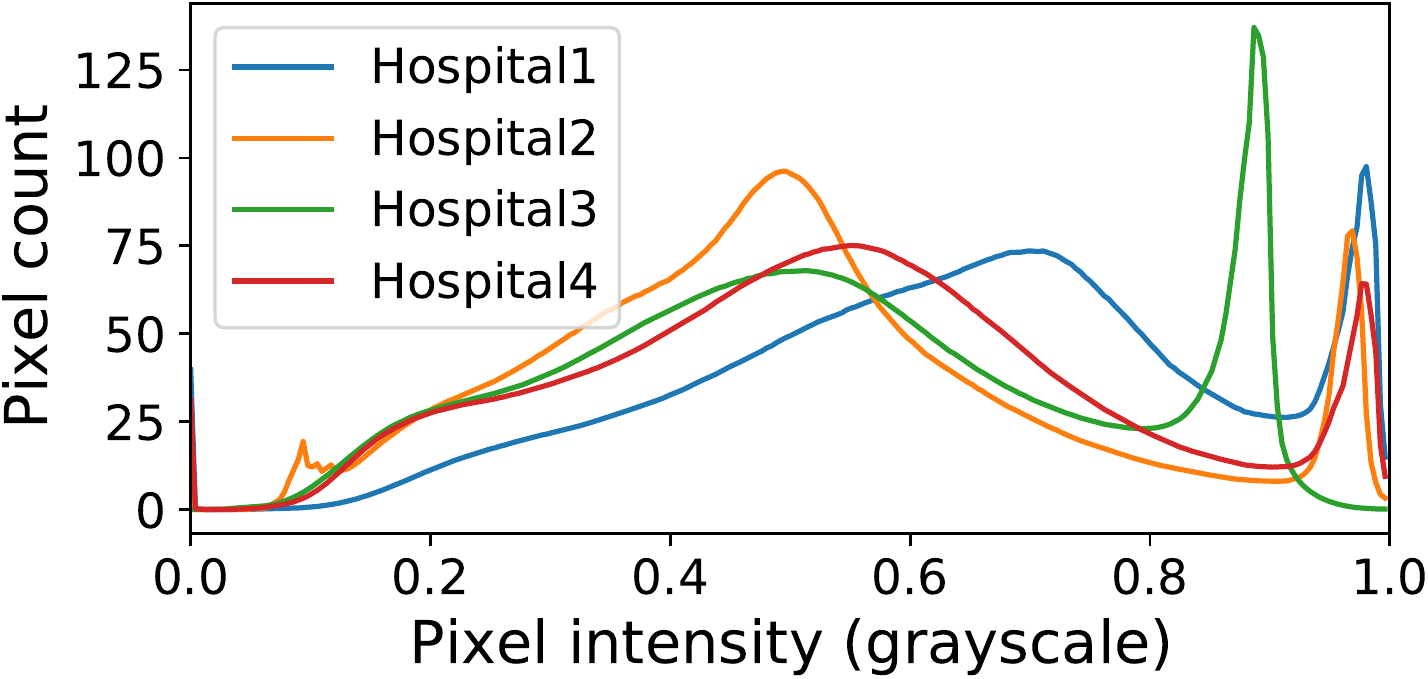}
        %  \vspace{-3mm}
         \caption{Camelyon17 distribution before CCST.}
         \label{fig:hist1}
     \end{subfigure}
     \hfill
     \begin{subfigure}[b]{0.49\textwidth}
         \centering
         \includegraphics[width=0.9\textwidth]{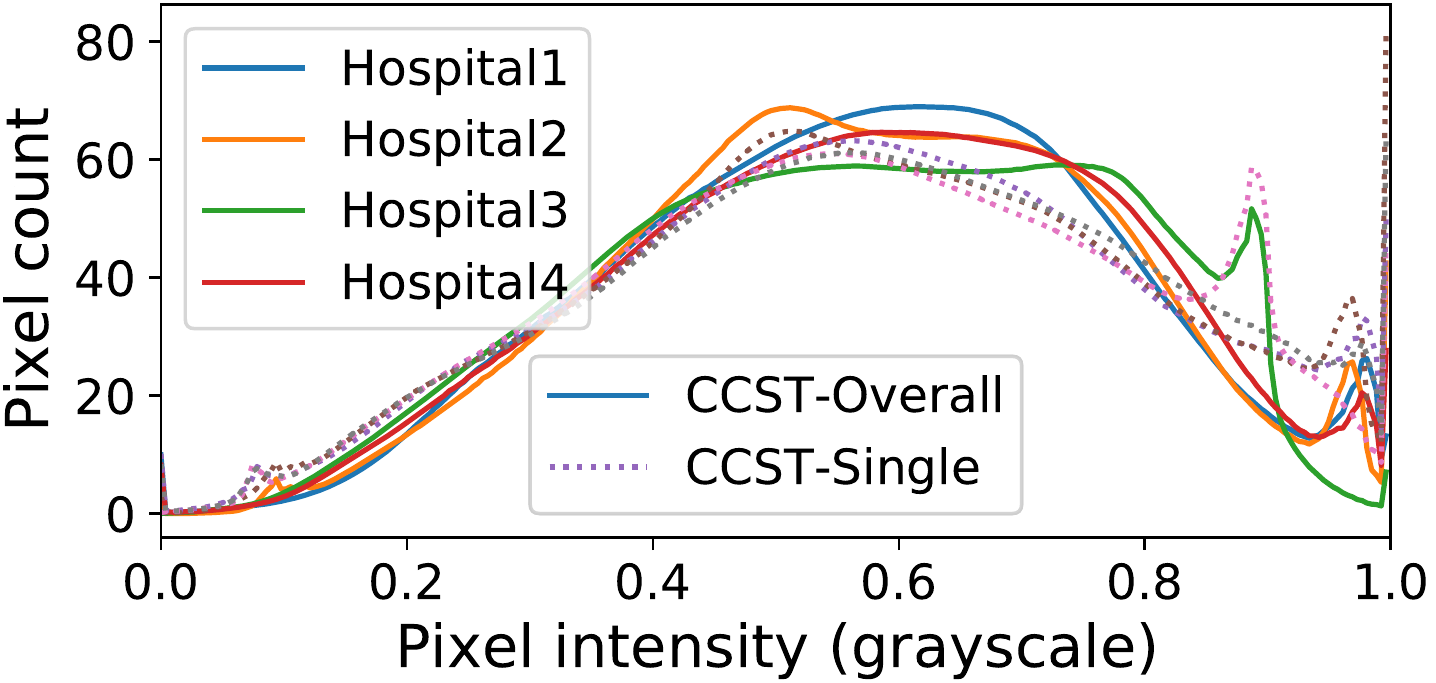}
        %  \vspace{-3mm}
         \caption{Camelyon17 distribution after CCST.}
         \label{fig:hist2}
     \end{subfigure}
        \caption{The distribution of source clients on the Camelyon17 before and after our CCST when the target client is hospital 5. The y-axis is the average pixel count per image, and the x-axis is the greyscale values. Note that we convert the RGB into grey images for distribution visualization. \textbf{(a)} Before CCST, the distributions of source clients are not uniform. \textbf{(b)} After CCST with either single image style or overall style, the distributions of source clients become much more uniform.}
        \label{fig:hist}
\end{figure*}

\subsection{Results}
\begin{figure*}[t]
     \centering
     \begin{subfigure}[b]{0.49\textwidth}
         \centering
         \includegraphics[width=0.9\textwidth]{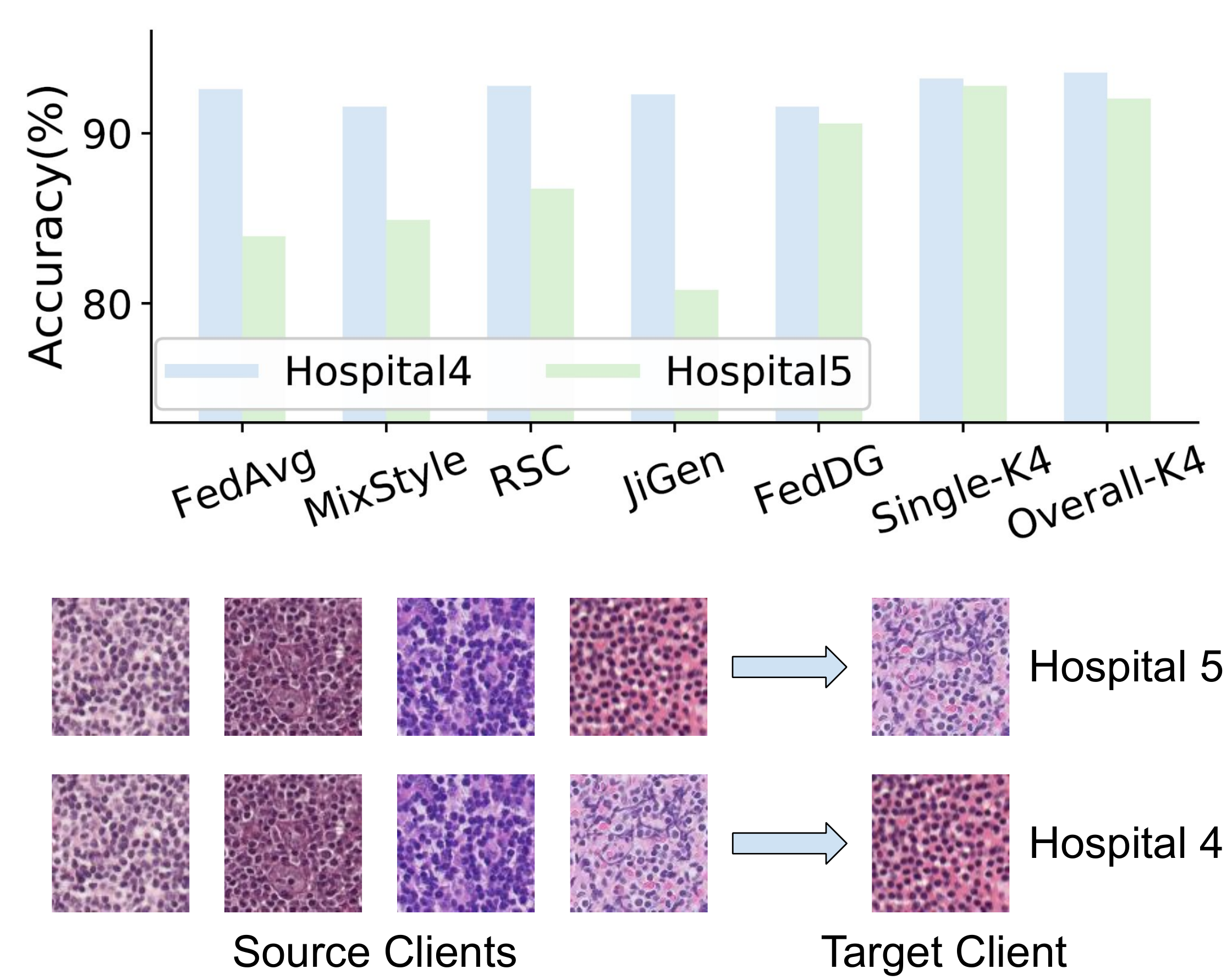}
        %  \vspace{-3mm}
         \caption{Results on the Camelyon17.\vspace{-5pt}}
         \label{fig:camelyon17_results}
     \end{subfigure}
     \hfill
     \begin{subfigure}[b]{0.49\textwidth}
         \centering
         \includegraphics[width=0.9\textwidth]{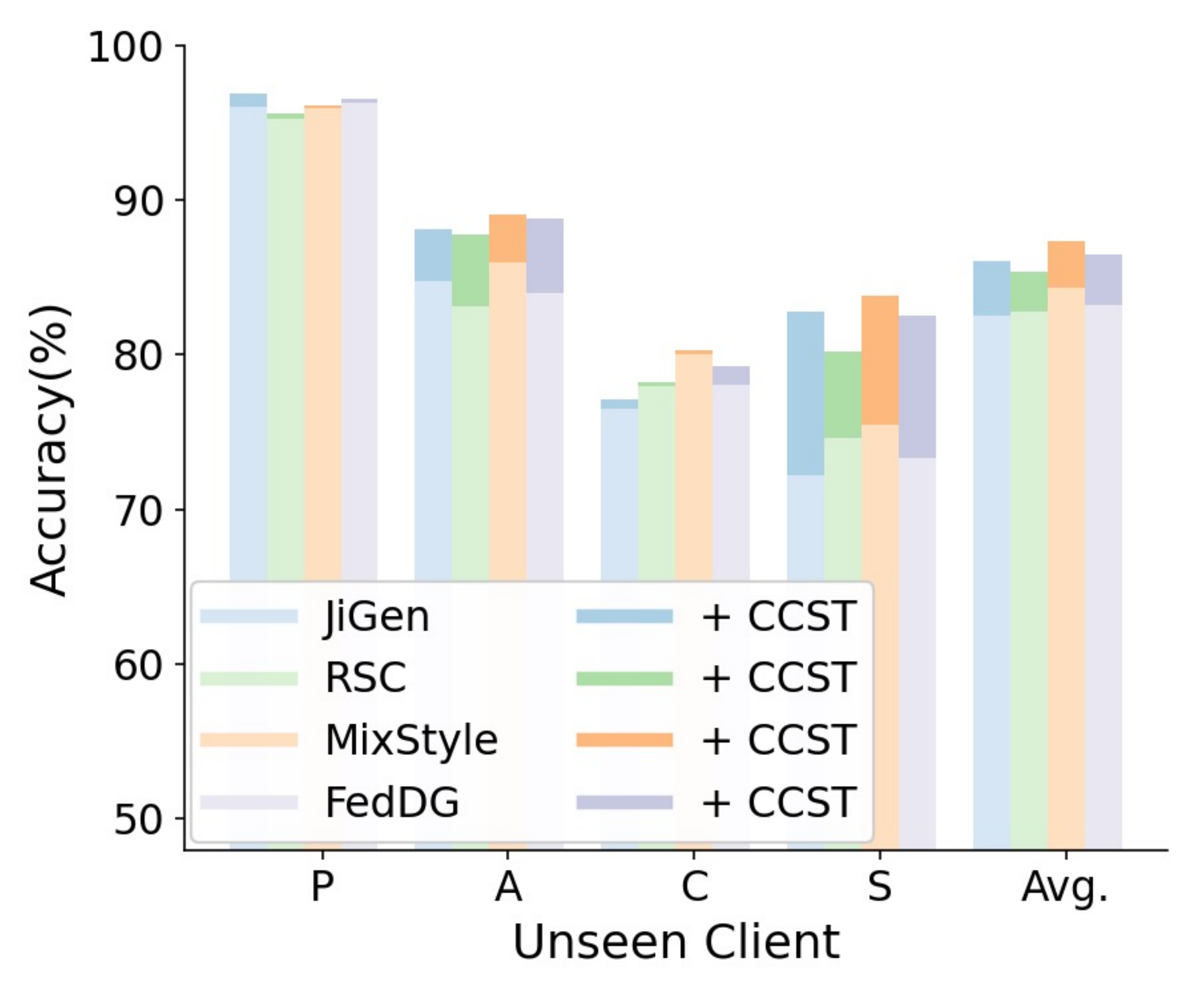}
        %  \vspace{-3mm}
         \caption{Orthogonality on the PACS.\vspace{-5pt}}
         \label{fig:extra_boost}
     \end{subfigure}
        \caption{\textbf{(a)} The results on the Camelyon17 dataset. Our method outperforms other DG methods when tested on hospitals 4 and 5. \textbf{(b)} Extra performance boost on other domain generalization methods with our cross-domain style transfer (CCST) with overall style on the PACS dataset. Each x-tick represents the single unseen client in a leave-one-client-out experiment, and Avg. is abbreviated for the average accuracy.\vspace{-13pt}}
        \label{fig:bars}
\end{figure*}

\textbf{Comparison with state-of-the-arts.} We compare our approach with three centralized DG methods and a federated DG method for on standard DG benchmarks PACS and OfficeHome as well as a real-world medical image dataset Camelyon17. 
Table~\ref{table:all_results} presents the quantitative results of the image recognition task for different target clients on both PACS and Office-Home datasets. Each single letter column shows the test accuracy of the global model with the best validation accuracy on an unseen client. 
Although all DG methods can have better performance based on FedAvg, our approach demonstrates a significant boost over others on both datasets. On the PACS benchmark, our method achieves the average accuracy of 86.52\% , which is 3.47\% better than the second best method FedDG. Especially for the unseen client S (Sketch), CCST outperforms other methods by more than 7\%. When photo is the target domain, all the methods perform similarly because we start training based on the ImageNet-pretrained model, which already has very high performance on photo images.
Besides the PACS benchmark, the performance of our approach on the Office-Home dataset also has consistent results. Specifically, CCST outperforms other methods on average with a testing accuracy of 63.56\%. 
Due to the small discrepancy in domain styles, all those domain generalization methods bring smaller improvements (less than 1\%) than on the PACS. Overall, CCST outperforms other DG methods by a large margin.
Figure~\ref{fig:camelyon17_results} shows the results on the Camelyon17 dataset, our method outperforms other DG methods both when hospital 4 and hospital 5 as the target client. Some DG method, such as JiGen, is even harmful when applied in FL setting on the Camelyon17 dataset.   

Our CCST enables each local client model to update under more diversified images, thus providing a wide range of image styles. The more diversified intra-client distribution helps reduce the bias towards fitting a specific distribution and update the model with a more general direction. As shown in Figure~\ref{fig:hist}, the distribution of source clients become much more uniform after applying our CCST method.
% making each local model has a unified goal to fit all the styles and avoiding averaging local models with different biases. 
In contrast, for domain generalization methods aim to learn more general features (e.g., Jigen, RSC), they suffer from limited intra-client distribution and different inter-client distribution, only achieving marginal improvements. 
Similar to CCST, MixStyle and FedDG also aim to increase the model generalization ability by diversifying the feature distribution. While in the federated setting, MixStyle can only perform intra-client feature diversification by assuming every single image has a unique style, which set a limit on its diversity level. FedDG utilizes amplitude information in frequency space of an image as a kind of style information, and perform amplitude exchange to diversify the data distribution across clients. Nevertheless, the amplitude exchange leads to minor appearance change, which may not enough for images with a large domain gap, while CCST utilizes style transfer to give a more thorough style exchange across clients, which leads to better results. 
We give visualization results after style transfer in the Section~\ref{sec:visualization} of the supplementary. The overall domain style usually represents a more general and accurate client style, while the single image style brings more randomness.

\begin{table*}[t]
    \caption{\textbf{(a)} Performance of our approach using ResNet50 as the backbone with four different image style transfer settings compared with the baseline of FedAvg on the PACS benchmark. Each column represents a single unseen target client. \textbf{(b)} Performance of our approach with test time adaptation (Tent)       \cite{wang2021tent} on the PACS benchmark using ResNet50.}
    \vspace{-5pt}
    \begin{subtable}[h]{0.5\textwidth}
        \centering
        \resizebox{\textwidth}{!}{
        \begin{tabular}{ccccccc}
\toprule
 & \multirow{2}{*}{Setting} & \multicolumn{4}{c}{Unseen client}                                         & \multirow{2}{*}{Average} \\ \cline{3-6}
                          &                          & P              & A             & C              & \multicolumn{1}{c}{S} &                          \\ \hline
& FedAvg~\cite{mcmahan2017communication} & 95.51          & 82.23          & 78.20          & 73.56                   & 82.37                    \\ \cline{2-7}
&  Single (K=1)           & 95.75&87.5&74.66&76.56&83.62\\
& Single (K=2) &\textbf{96.77}&86.23&75.73&80.12&84.71\\
& Single (K=3)            & 96.65&86.63&74.53&81.85&84.84\\
& Overall (K=1)           & 95.69&86.67&75.85&77.37&83.90\\
&Overall (K=2) &96.41&\textbf{88.72}&78.03&80.91&86.02\\
& Overall (K=3)           & 96.65&88.33&\textbf{78.20}&\textbf{82.90}&\textbf{86.52}\\\hline            
\end{tabular}}
       \caption{Control experiment for CCST. \vspace{-20pt}}
       
       \label{tab:ablation}
    \end{subtable}
    \hfill
    \begin{subtable}[h]{0.49\textwidth}
        \centering
        \resizebox{\textwidth}{!}{
        \begin{tabular}{cccccc}
\toprule
\multirow{2}{*}{Setting} & \multicolumn{4}{c}{Unseen client}                                         & \multirow{2}{*}{Average} \\ \cline{2-5}
                                              & P              & A             & C              & \multicolumn{1}{c}{S} &                          \\ \hline
EoA~\cite{arpit2021ensembledg} & {98.00} & {90.50} & {83.40} & {82.50} & {88.60} \\\hline
Single (K=1) & 97.78& 89.55& 84.51& 82.79& 88.66 \\
Single (K=2) &97.54& 89.40& 84.43& 85.42& 89.20\\
Single (K=3) & 98.08& 89.75& 86.05& 86.03& 89.98 \\
Overall (K=1) & 97.78& \textbf{90.92}& 86.01& 83.66& 89.59 \\
Overall (K=2) & \textbf{98.38}& 90.72& 86.47& 85.62& 90.30 \\
Overall (K=3)  & 98.14& 90.87& \textbf{86.77}& \textbf{86.10}& \textbf{90.47}  \\ \hline
\end{tabular}}
        \caption{CCST+Tent~\cite{wang2021tent}. \vspace{-20pt}}

        \label{tab:add_tent}
     \end{subtable}
     
     \label{tab:new_table}
\end{table*}

% \subsection{Control Experiments on CCST}
\textbf{Control experiments on CCST.}
We conduct control experiments to investigate two types of image style with different augmentation levels. In Table \ref{tab:ablation}, \textit{Single} and \textit{Overall} represents single image style and overall domain style mentioned in Section \ref{sec:method} respectively. Different augmentation level $K$ indicates the intensity of augmentation.  We evaluate the four settings on the PACS benchmark with ResNet50. 
%From Table~\ref{table:ablation} we can see that Overall ($K=2$) can already achieve promising improvements. For the Overall ($K=3$), it obtains a similar result with $K=2$, and we apply this setting when compared with other methods. 
For each kind of style, larger augmentation level K leads to a better performance. It is worth mentioning that the performance achieved by K=2 is similar with K=3, which indicates that our method can already achieve good performance with a relatively large K. 
For different types of style, overall domain style shows more improvement than single image style because the overall style is able to represent a more general and accurate domain statistics, while single image styles may differ a lot due to randomness.

% \subsection{Orthogonality}
\textbf{Orthogonality.}
Our method is orthogonal to many other DG methods and can lead to additive performance via combined utilization. As many traditional domain generalization methods require centralized data and need to make use of various styles to achieve a domain robust model, CCST can serve as an initial step to benefit traditional DG methods with diversified styles. As shown in Figure~\ref{fig:extra_boost}, we plot the average test accuracy of FedDG and three centralized DG methods on the PACS benchmark before and after applying our CCST. From the average accuracy, we can see all DG methods benefit a further boost with the help of cross-client style transfer (CCST). Interestingly, we find the performance when tested on sketch client (S) gains the largest improvement with CCST.
Besides, we also extend our method with Tent~\cite{wang2021tent} which uses entropy to update parameters in batch normalization layers at test time. As shown in Table~\ref{tab:add_tent}, with this method combined in federated setting, our method surpasses the state-of-the-art DG method EoA~\cite{arpit2021ensembledg} on the PACS benchmark in centralized setting with an average test accuracy of 90.47\%.

\subsection{Discussions} 
\label{sec:discussion}
\textbf{Computation and communication trade-off.} To reduce the communication cost that is usually known as the bottleneck of FL, we make a trade-off of using extra local computational cost in the overall style computation stage. Specifically, for the overall style, it requires a lot of local computation. A possible solution could be only choosing a relatively large portion of images from each class to approximate the overall domain style. After local style computation, only lightweight style vectors (two 512 dimensional vectors for each client) will be communicated between local clients and the central server for once.
The extra computation cost of our CCST method is very low, we give a quantitative analysis in the Section \ref{sec:time_cost} of the supplementary.

% \textbf{Analysis on bad cases.} Although CCST has outperformed other methods by introducing more diverse distributions for each client, transferring the style of context-rich images (e.g., photo) to content-poor images (e.g., sketch) may not lead to good results. This is because the structure of content-poor images is too simple and almost has no context differences (e.g., pure white background in Sketch images). It could be an interesting future work direction to explore how to introduce prior guidance for transferring rich styles to content-poor images with few contexts.

\textbf{Analysis for the privacy issue of the shared style vector.} Like all federated learning methods, our method may also suffer from privacy issue. However, it is difficult to reconstruct the original datasets with only a 1024-dimensional style vector shared. First, different datasets can have the same style vector. Second, the style vector is the statistics of the pixel feature set, which has no order. Even if we assume that the original pixel feature set can be reconstructed, it is hard to rearrange those pixels into proper order. This is even more difficult for the overall style vector because it is the set statistics of pixels features from many images. Therefore, one can hardly recover the whole datasets only with our shared style vectors. We further experimented with reconstructing the original images from style vectors using a state-of-the-art GAN generator~\cite{liu2021towards}. The experiment shows that neither can a malicious source client recover other clients' images only from style vectors, nor an outside attacker can reconstruct source images using a pre-trained generator. Please refer to the Section~\ref{sec:privacy} in the supplementary for more details.
% \begin{enumerate}
% \item  CCST may not transfer simple images (e.g., sketch) into a realistic context-rich images.
% \item  line 496 (computatino cost)
% \item  computation
% \end{enumerate}
\section{Conclusion}

In this paper, we propose a novel DG method for image recognition under the FL setting. Our method utilizes cross-client style transfer to introduce all the source client styles into each client while not violating the original data-sharing restriction of FL. Two types of styles and corresponding sharing mechanisms are proposed to be utilized accordingly. The state-of-the-art generalization performance of our method under the FL setting is demonstrated on two standard DG datasets and a real-world medical image dataset with comprehensive experiments. The data style transfer strategy opens a new way for heterogeneous clients to access diversified distributions without sharing original data, which helps the global model generalize better in FL. Moreover, our method is orthogonal to many other SOTA DG methods and can be combined together to have further performance boost.

%-------------------------------------------------------------------------
{\small
\bibliographystyle{ieee_fullname}
\bibliography{ref}
\balance
}

\clearpage
\appendix

\begin{figure*}[t]
     \centering
     \begin{subfigure}[b]{0.49\textwidth}
         \centering
         \includegraphics[width=0.99\textwidth]{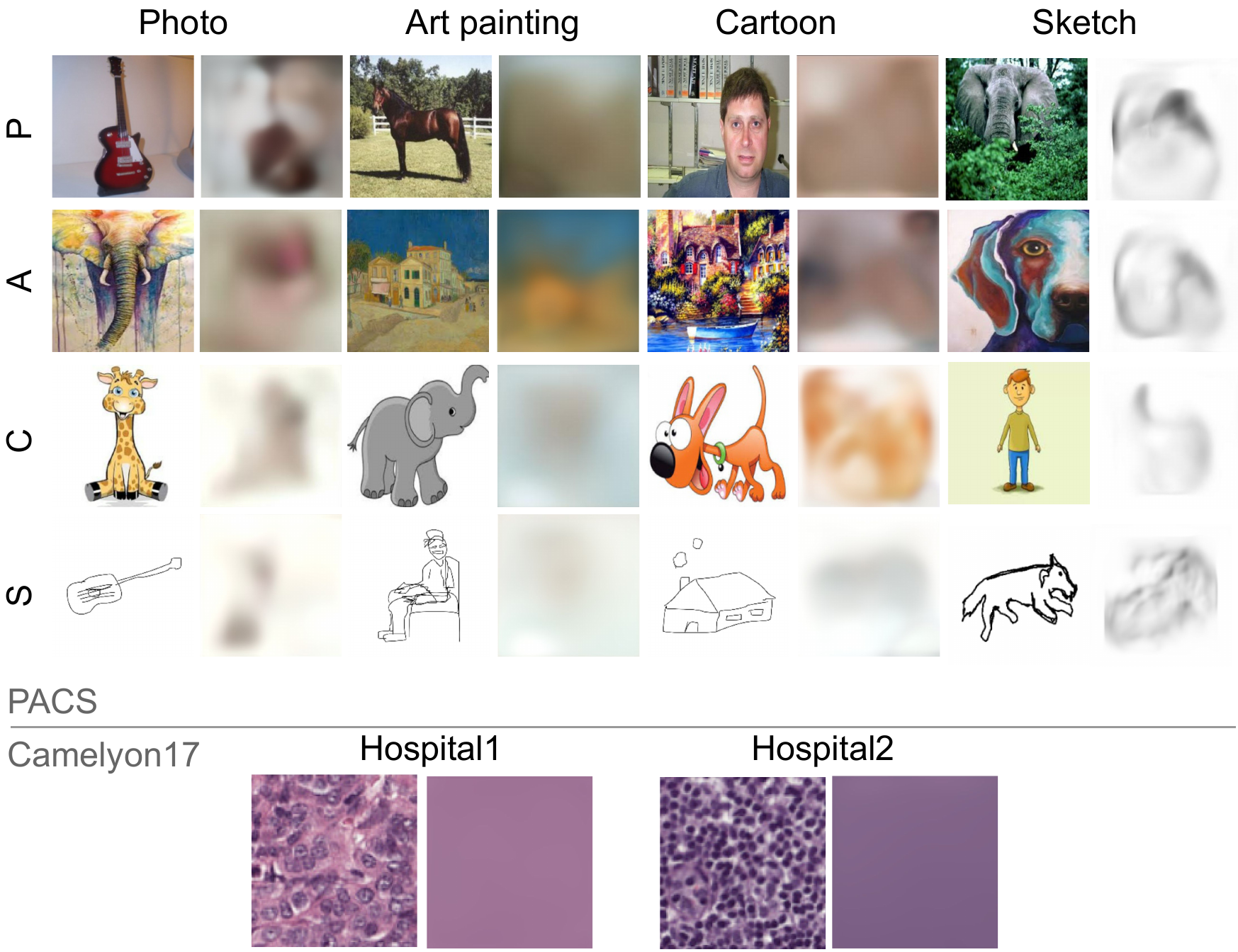}
        %  \vspace{-3mm}
         \caption{Intra and inter domain reconstruction.}
         \label{fig:recons_intra_inter}
     \end{subfigure}
     \hfill
     \begin{subfigure}[b]{0.49\textwidth}
         \centering
         \includegraphics[width=0.99\textwidth]{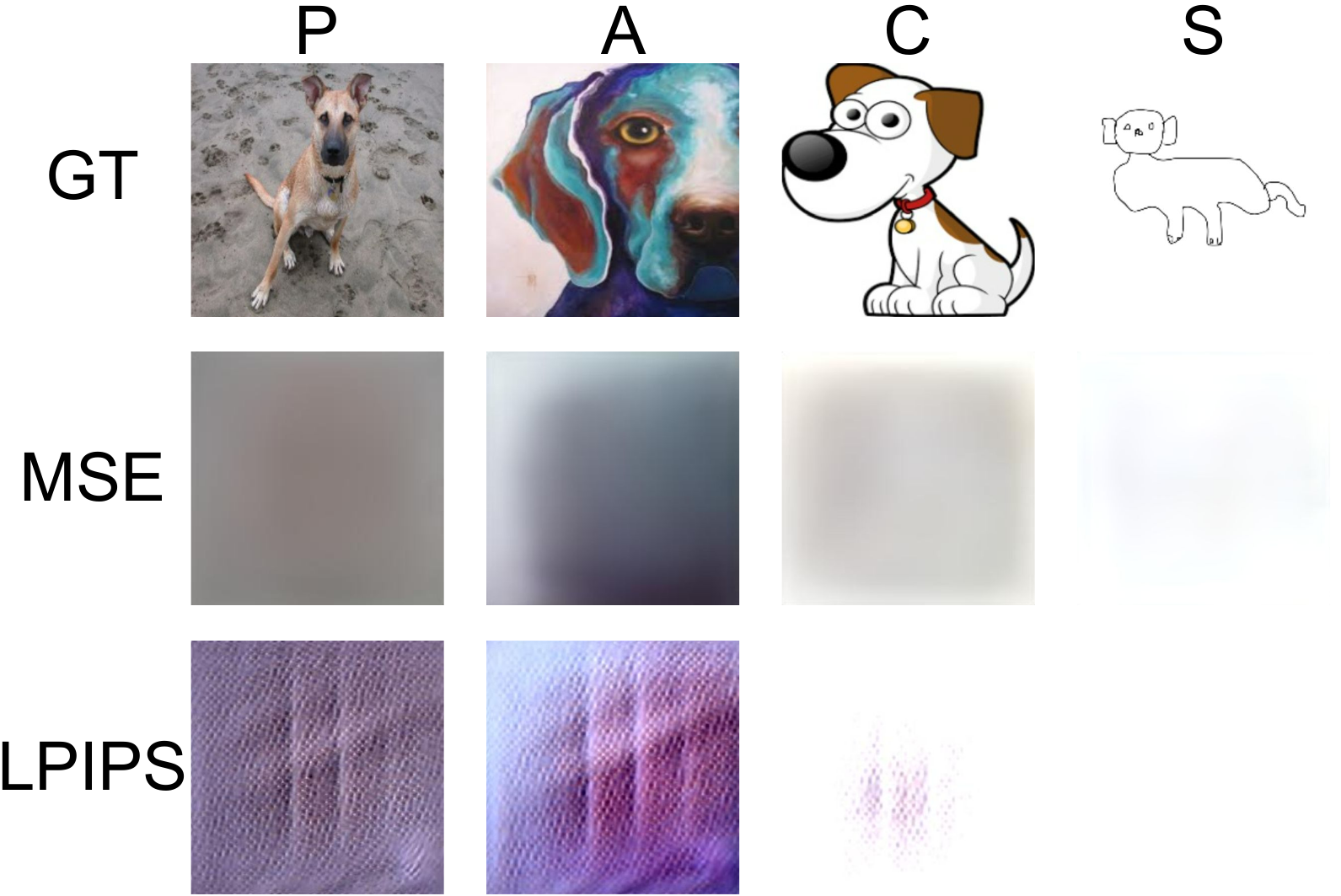}
        %  \vspace{-3mm}
         \caption{ImageNet pre-trained reconstruction.}
         \label{fig:recons_imagenet}
     \end{subfigure}
        \caption{\textbf{(a)} Image reconstruction from the style vectors. (Ground truth, reconstructed image) pairs are shown. For PACS, the x-axis represents which domain the generator is trained on, the y-axis represents which domain the input style vectors are from. Diagonal image pairs show intra-client results. For Camelyon17, we show the intra-client reconstruction results only. The generator is trained with MSE loss. \textbf{(b)} We utilize the ImageNet pre-trained reconstructor to recover the PACS images from their style vectors. From top to bottom rows are the ground truth images, reconstruction results using generator trained with MSE loss, and reconstruction results using generator trained with LPIPS loss. }
        \label{fig:recons}
\end{figure*}

% \begin{figure}[h]
% \centering
% \includegraphics[width=0.99\columnwidth]{pics/DG4FL_rebuttal.pdf} 
% \caption{}
% \label{fig:recons}
% \end{figure}

\section{Image reconstruction from style vectors}
\label{sec:privacy}
To evaluate the safety of the style vectors, we train a generator from a SOTA GAN~\cite{liu2021towards} to reconstruct the image from its style vector. We train the generator until the validation loss converges sufficiently. The best model is selected with the highest validation average PSNR. The results are shown in Figure \ref{fig:recons}. We consider three scenarios: 
\begin{itemize}
\item First, intra-client reconstruction (train and test on data from the same client). Note that this scenario is impossible in FL unless the client has already leaked their data. This is an extreme case to examine the best results the generator can achieve. 

In Figure \ref{fig:recons_intra_inter}, the diagonal image pairs show intra-client reconstruction results on PACS, and the bottom two pairs show intra-client reconstruction results on Camelyon17. We can see that the generator fails on intra-client reconstruction on the Camelyon17 dataset. For PACS, although the generator is possible to overfit the data within a single domain, this is only vulnerable when a large amount of data is leaked.

\item Second, inter-client reconstruction (malicious client). It is possible that there exists a malicious client who wants to use its own data to reconstruct the images of other clients from the shared style vectors. From the results, we can hardly infer any content information except for overall color. Although (P, Photo) shows a rough shape of GT, it belongs to the intra-client scenario, which violates the FL setting. 

In Figure \ref{fig:recons_intra_inter}, image pairs that are not lie at diagonal line are inter-client reconstruction results. The figure shows that the inter-client reconstruction fails on both PACS and Camelyon17 datasets.

\item Third, third-party reconstruction (pre-trained on large-scale images). More generally, if an outside attacker has compromised the style vectors and wants to reconstruct the images from the shared style vectors, they can train the reconstructor on a large-scale image dataset. 

We train the generator on ImageNet and visualize the reconstruction results in Figure \ref{fig:recons_imagenet}. According to the results, the pre-trained generator totally fails to reconstruct the target images. 

\end{itemize}

Therefore, in the real FL scenarios, one can hardly reconstruct the original images merely from the shared style vectors.

\section{Time cost of extra computation}
\label{sec:time_cost}
The extra computation time cost of our method is very low. Specifically, for the overall style computation, it takes 7 seconds for 2048 images with 256$\times$256 resolution. For image stylization, it takes 54 seconds to stylize 2048 images of 256$\times$256 resolution under either “Overall, K=3” or “Single, K=3” mode. 
The results are tested on an NVIDIA RTX 2080Ti GPU using PyTorch 1.11.0 with CUDA11.

\begin{figure*}[t]
\centering
\includegraphics[width=0.9\textwidth]{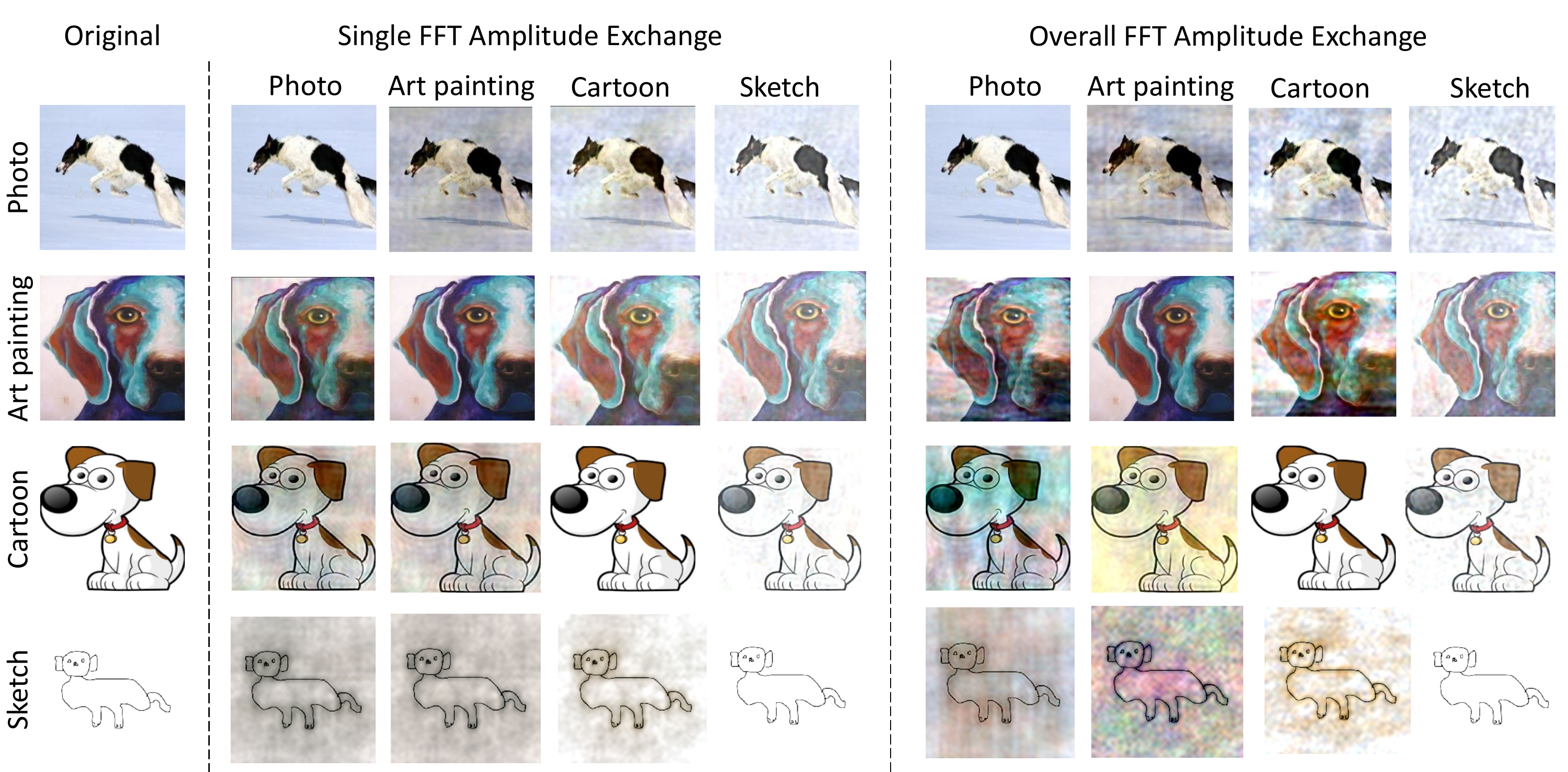} 
\caption{Visualization of images after the FFT amplitude exchange on the PACS dataset. Similar with Figure \ref{fig:visual_more}, we duplicate the content image if it is the same as the amplitude target image.}
\label{fig:fft_compare}
\end{figure*}

\section{Training budget}
To be fairer in the training budget, we increase the local training iterations of baselines methods from 1 to 3 to compare with our overall (K=3) method. The results are shown in Table \ref{tab:budget}. According to the results, more local iterations do not lead to obvious accuracy improvement for baseline methods, and our CCST (Overall, K=3) still outperforms all the baseline methods.

 \begin{table}[t]
        \centering
        \resizebox{0.5\textwidth}{!}{
        \begin{tabular}{ccccccc}
\toprule
 & \multirow{2}{*}{Setting} & \multicolumn{4}{c}{Unseen client}                                         & \multirow{2}{*}{Average} \\ \cline{3-6}
                          &                          & P              & A             & C              & \multicolumn{1}{c}{S} &                          \\ \hline
& FedAvg (AISTATS'17)~\cite{mcmahan2017communication}    &95.21&82.91&78.80&73.99&82.73 \\
& Jigen (CVPR'19)~\cite{carlucci2019domain}         &95.63&83.25&81.10&71.95&82.98 \\
& RSC (ECCV'20)~\cite{huang2020self}           &94.55&83.20&79.99&72.79&85.31 \\
& MixStyle (ICLR'21)~\cite{zhou2021domain}           &96.47&86.89&81.06&76.81&82.63 \\
& FedDG (CVPR'21)~\cite{liu2021feddg}  &95.93&84.28&79.44&73.89&83.89 \\
& CCST (Overall,K=3)           & \textbf{96.65}&\textbf{88.33}&\textbf{78.20}&\textbf{82.90}&\textbf{86.52}\\\hline            
\end{tabular}}
       \caption{Compare the results of our CCST (Overall, K=3) with baselines that are trained with local iterations=3.}
       \vspace{-3mm}
       \label{tab:budget}
    \end{table}

% FedAvg (AISTATS'17)~\cite{mcmahan2017communication}& 95.51& 82.23& 78.20& 73.56& 82.37 
% & 60.08 & 45.59 & 69.48& 72.82& 61.99&72.18\\
% Jigen (CVPR'19)~\cite{carlucci2019domain}& 95.99& 84.72& 77.09& 72.16  & {82.49} 
% &  60.29 & 46.16 & 69.26 & 72.59& 62.07&72.28\\
% RSC (ECCV'20)~\cite{huang2020self}& 95.21& 83.15& 78.24& 74.62&{82.81}
% & 58.23 & 46.05  & 70.27 & \textbf{73.39}&61.99&72.40\\
% MixStyle (ICLR'21)~\cite{zhou2021domain}& 95.93& 85.99& \textbf{80.03}& 75.46 & {84.35}
% & 58.44 & \textbf{50.29} & {70.61} & 70.64 & 62.49&73.42\\
% FedDG (CVPR'21)~\cite{liu2021feddg}& 96.23& 83.94& 79.27& 73.30& {83.19} 
%  & \textbf{60.70} & 45.82 & {71.51} & {73.05}& 62.77&72.98\\
% CCST (Overall, K=3) & \textbf{96.65}& \textbf{88.33}& 78.20 & \textbf{82.90} & \textbf{86.52}
%  & {59.05} & {50.06} & \textbf{72.97}& {71.67}& \textbf{63.56} & \textbf{75.04} \\ \bottomrule
% \end{tabular}

\section{Visualization of the FFT amplitude exchange results on the PACS}

As shown in Figure~\ref{fig:fft_compare}, we visualize the results after the FFT amplitude exchange using single and overall amplitude on the PACS dataset~\cite{PACS}. We can see that the FFT amplitude exchange does not make a noticeable change to appearance or artistic style but only adds to some spatial repetitive texture and color patterns. This could be one of the reasons why FFT cannot outperform our CCST method. Because in the PACS dataset, we have large domain gaps such as that between photos and sketches. Simple changes in color, brightness, or background texture cannot make up the gap very well, while AdaIN style transfer can perform better by producing visually plausible artistic style transfer.

\section{AdaIN style transfer vs FFT amplitude exchange}
\begin{table*}[t]
\centering
\caption{Comparison between using FFT amplitude and IN statistics as style in our proposed cross-client style transfer framework with single and overall style under different K values. This table reports the performance with varying hyper-parameters of our framework on the PACS dataset with ResNet50 as the backbone. }
\begin{tabular}{cccccc|ccccc}
\toprule
\multirow{4}{*}{Method} & \multicolumn{5}{c}{CCST (Ours)}                                & \multicolumn{5}{c}{FFT Amplitude Exchange}                         \\ \cmidrule(lr){2-11} 
% & \multicolumn{5}{c}{ResNet-50 (pre-trained on ImageNet)} & \multicolumn{5}{c}{ResNet-18 (pre-trained on ImageNet)} &        \\\cmidrule(lr){2-11}
& \multicolumn{1}{c}{P}         & A         & C         & S         & \multicolumn{1}{c}{Avg.}     & P         & A         & C        & S        & \multicolumn{1}{c}{Avg.}    \\ \midrule
Single(K=1)  & 95.75&87.5&74.66&76.56&83.62
&96.71&85.69&76.19&73.76&83.09 \\
Single(K=2)  &\textbf{96.77}&86.23&75.73&80.12&84.71
&96.89&87.16&79.31&74.32&84.42\\
Single(K=3)   & 96.65&86.63&74.53&81.85&84.84
&96.65&\textbf{86.87}&\textbf{79.74}&77.3&\textbf{85.14}\\
Overall(K=1)  & 95.69&86.67&75.85&77.37&83.90
&96.89&86.38&78.92&72.36&83.64\\
Overall(K=2)  &96.41&\textbf{88.72}&78.03&80.91&86.02
&93.95&79.79&72.18&77.96&80.97\\ 
Overall(K=3)  & 96.65&88.33&\textbf{78.20}&\textbf{82.90}&\textbf{86.52}
&95.21&81.25&73.34&\textbf{80.27}&82.52\\ \bottomrule
\end{tabular}
\label{table:adain-fft}
\end{table*}

In FedDG~\cite{liu2021feddg}, the amplitude information in the frequency space of an image can also serve as a kind of style, while we utilize the IN statistics of each feature channel as style information. To explore the differences between the FFT~\cite{nussbaumer1981fast} amplitude and IN statistics as style, we made a thorough comparison under our augmentation framework. The amplitude exchange alone without episodic learning in FedDG is equivalent to our framework with the setting of the single style when K=1.

We show the comparison results in Table~\ref{table:adain-fft}. Compared with our proposed method (using IN statistics as style), the FFT-based amplitude exchange method consistently performs worse under the same setting except for the setting of single style when K=3. Moreover, for the FFT-based style, the overall amplitude~\footnote{We compute the overall amplitude by averaging the amplitudes of all images in this domain.} of a domain fails to result in better results than a single image FFT amplitude. In contrast, our framework has a significant boost when using overall style. With the help of our framework, the best result (single, k=3) of the FFT-based method can have a 2\% improvement compared with the original version in FedDG (single, K=1). In general, our second-best result (overall, K=2) outperforms the best result of FFT amplitude exchange (single, K=3) by 0.9\%; our best result (overall, K=3) outperforms the best result of FFT amplitude exchange (single, K=3) by 1.4\%.

The experiments show that it is only practical to use the single image amplitude for the FFT amplitude exchange method. Utilizing the single image style mode makes the communication cost high due to the uploading and downloading of the style bank, leading to inflexibility. However, our CCST method can flexibly choose between single image style and overall domain style accordingly, especially the choice of using overall domain style to decrease the communication cost.

\begin{figure*}[t]
\centering
\includegraphics[width=0.95\textwidth]{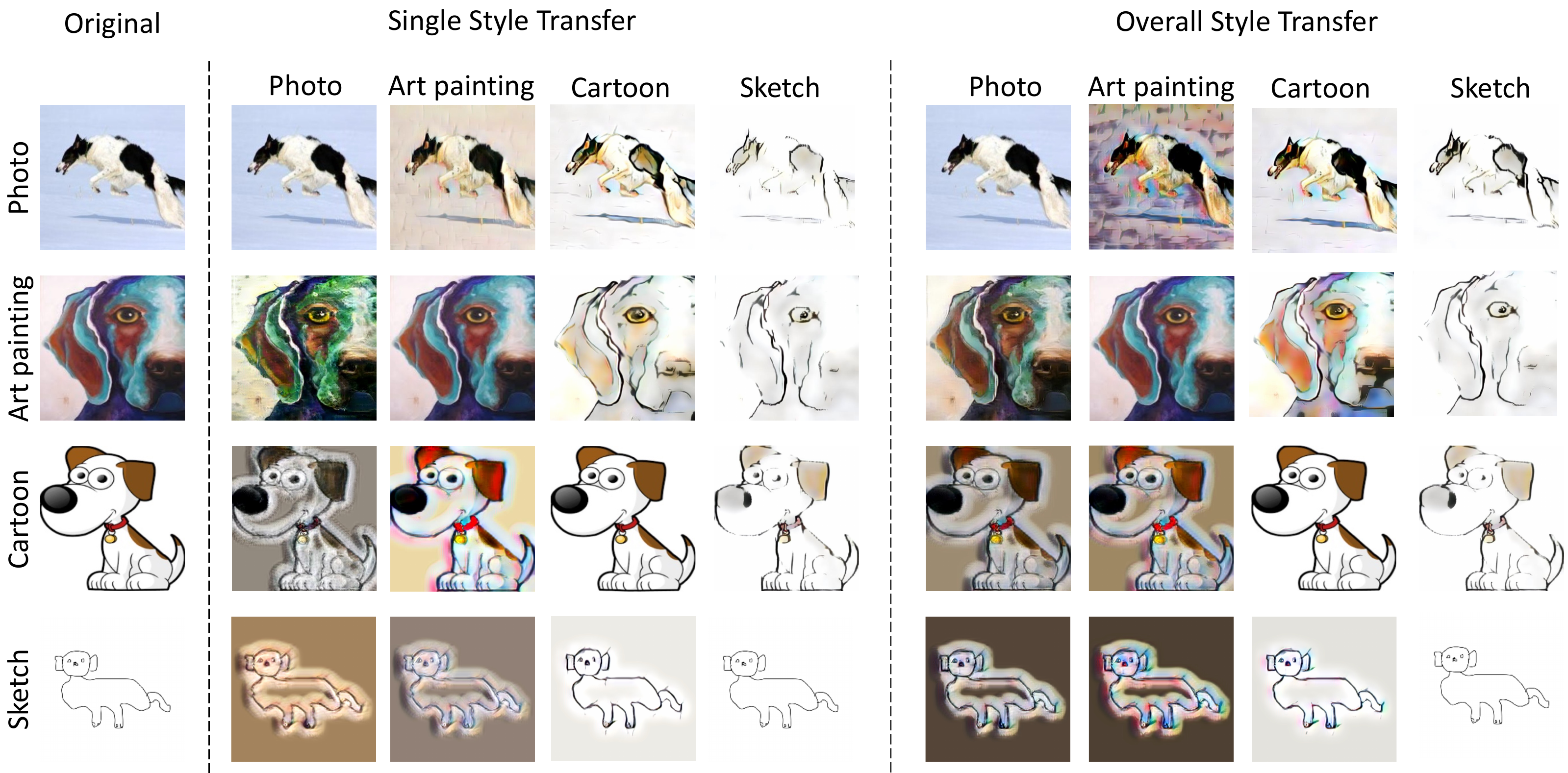} 
\vspace{-3mm}
\caption{Visualization of stylized images on PACS. Note that if the content image is from the same domain of the style statistics, we directly copy the content image to the augmented dataset instead of transferring the same style to it.}
\label{fig:visual_more}
\end{figure*}

\section{PACS visualization}
\label{sec:visualization}
Figure~\ref{fig:visual_more} shows the visual results of cross-client style transfer with two types of styles. The overall domain style represents a more general and accurate client style, while the single image style brings more randomness.

\section{Visualization of style transfer results on the Office-Home}
\begin{figure*}[t]
\centering
\includegraphics[width=0.9\textwidth]{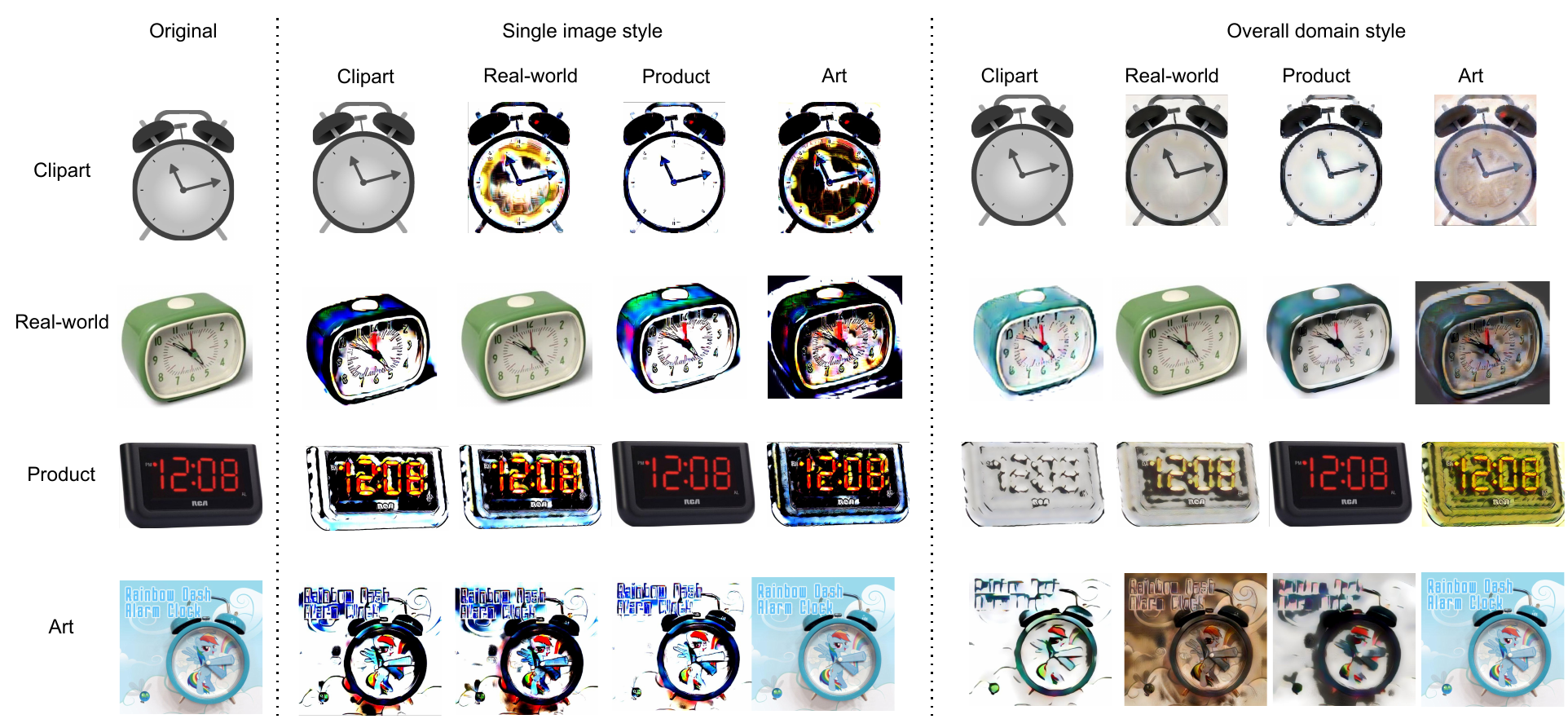} 
\caption{Visualization of the AdaIN stylized images on the OfficeHome dataset. Note that if the content image is from the same domain as that of style statistics, we directly copy the content image to the augmented dataset instead of transferring the same style to it.}
\label{fig:officehome_style_transfer}
\end{figure*}

In this section, we show the qualitative results by visualizing images before and after the AdaIN~\cite{huang2017adain}-based style transfer. In Figure~\ref{fig:officehome_style_transfer}, we show images of four different target domains in the Office-Home dataset~\cite{officehome}. Except for the art domain, samples from the other three domains show less domain gap. For each domain, we visualize the generated images using both random single image style and overall domain style. According to our experiment results, the overall style is usually more effective than using the single image style. Random single image style sometimes may choose an image that is not representative for the whole domain. For example, in Figure~\ref{fig:officehome_style_transfer}, when transferring the clock image with the Clipart style into real-world style, the stylized image with overall style has a more colorful and representative style than that using random single image style.

\begin{table*}[t]
\centering
\caption{Results of our CCST with different image style types and K values under PACS and Office-Home dataset. The backbone network is ResNet18. Each column represents a single unseen target client. }
\begin{tabular}{cccccc|ccccc}
\toprule
\multirow{4}{*}{Method} & \multicolumn{5}{c}{PACS}                                & \multicolumn{5}{c}{Office-Home}                         \\ \cmidrule(lr){2-11} 
% & \multicolumn{5}{c}{ResNet-50 (pre-trained on ImageNet)} & \multicolumn{5}{c}{ResNet-18 (pre-trained on ImageNet)} &        \\\cmidrule(lr){2-11}
& \multicolumn{1}{c}{P}         & A         & C         & S         & \multicolumn{1}{c}{Avg.}     & A         & C         & P        & R       & \multicolumn{1}{c}{Avg.}    \\ \midrule

FedAvg~\cite{mcmahan2017communication}& 91.44&75.98&73.21&61.08&75.43
& 60.08 & 45.59 & 69.48& \textbf{72.82}& 61.99\\
Single(K=1)  &94.07&77.73&70.99&72.82&78.90
&55.14&43.64&68.58&68.92&59.07 \\
Single(K=2)  &\textbf{95.27}&79.05&72.82&77.88&81.26
& 57.61&48.68&71.17&71.44&62.23\\
Single(K=3)  &94.79&80.27&71.72&\textbf{80.86}&81.91
&58.44&45.70&72.30&71.56&62.00\\
Overall(K=1)  &94.19&79.88&72.14&75.41&80.41
& 59.47&47.88&67.91&70.87&61.53\\
Overall(K=2)  &93.95&79.79&72.18&77.96&80.97
&57.82&\textbf{50.52}&71.28&70.99&62.65\\ 
Overall(K=3)  &95.21&\textbf{81.25}&\textbf{73.34}&80.27&\textbf{82.52}
&{59.05}&50.06&\textbf{72.97}&{71.67}&\textbf{63.44}\\ \bottomrule
\end{tabular}
\label{table:res18}
\end{table*}

\section{Additional experimental results}

We show the results of our CCST with ResNet~\cite{he2016deep} as the backbone network on the PACS and Office-Home dataset in Table~\ref{table:res18}. For PACS dataset, using ResNet18 (Table~\ref{table:res18}) and ResNet50 (Table 2a) as backbone have consistent results: the overall style with K=3 leads to the best performance. When using ResNet18 as the backbone, the improvement upon baseline is more significant than that of using ResNet50. 

For CCST results on the Office-Home dataset, besides $K=1$, all other settings outperform the FedAvg in terms of the average accuracy. To explore the reason for the failure of CCST with $K=1$,  we visualize the images with style transfer as shown in Figure~\ref{fig:officehome_style_transfer}. From the visualization results, we can observe that the domain shift among domains of the Office-Home is smaller than that of the PACS dataset. For example, the product images collected on websites are similar to the real-world object images taken by a regular camera. Due to the slight differences between different domain styles and the randomness in single image style transfer, Single(K=1) achieves a lower accuracy than FedAvg on average. However, the overall domain style still shows a stronger representation capability and only has a minor performance gap compared with FedAvg. Overall, our best CCST results outperform the FedAvg baseline even with less domain shift in the Office-Home dataset.

% \\$[$\textcolor{green}{R2}$]$ \textbf{Unfair training budget.}
% We observed that training 500 epochs will let the training converge sufficiently, which means training more epochs after 500 will have little performance improvement. That is why we train 500 epochs.
% Therefore, even if it is unfair in training budget, the training already achieve their best.
% \begin{figure}[h]
% \centering
% \includegraphics[width=0.8\columnwidth]{pics/FedAdaIN.pdf} 
% \caption{placeholder}
% \label{fig:adain}
% \end{figure}

% \\$[$\textcolor{blue}{R4}$]$ \textbf{The value of M and whether it is online augmentation.}
% The value of M in L327-328 is $\lceil dataset\_size / 32\rceil$ in our experiment. Setting M to be the size of whole dataset results in similar results. The augmentation is done offline as described in L123-L125. Note that given the randomness of single style mode, it may also outperform the overall mode (e.g., the Single-K4 v.s. Overall-K4 in Figure 4a.)

\end{document}